\documentclass[pmlr]{jmlr}

\usepackage{longtable}

\usepackage{booktabs}
\usepackage[load-configurations=version-1]{siunitx} 
\usepackage[flushleft]{threeparttable}
\usepackage{multirow}

\theorembodyfont{\upshape}
\theoremheaderfont{\scshape}
\theorempostheader{:}
\theoremsep{\newline}

\jmlrvolume{XX}
\jmlryear{2019}
\jmlrworkshop{Machine Learning for Health (ML4H) at NeurIPS 2019}

\title[Improving the generalizability of automated medical abbreviation disambiguation]{Training without training data: Improving the generalizability of automated medical abbreviation disambiguation*}

  \author{\Name{Marta Skreta}\textsuperscript{1,2} \Email{martaskreta@cs.toronto.edu}\\
     \Name{Aryan Arbabi}\textsuperscript{1,2} \Email{arbabi@cs.toronto.edu}\\
   \Name{Jixuan Wang}\textsuperscript{1,2,3} \Email{jixuan@cs.toronto.edu}\\
   \Name{Michael Brudno}\textsuperscript{1,2} \Email{brudno@cs.toronto.edu}\\
   \addr \textsuperscript{1}University of Toronto, Department of Computer Science \\ \textsuperscript{2}The Hospital for Sick Children, Center for Computational Medicine
   \\ \textsuperscript{3}Vector Institute for Artifical Intelligence, Toronto, Canada
   }

\begin{document}
\thanks{No doctors were hurt annotating gold standard data for this paper.}
\maketitle

\begin{abstract}
Abbreviation disambiguation is important for automated clinical note processing due to the frequent use of abbreviations in clinical settings. Current models for automated abbreviation disambiguation are restricted by the scarcity and imbalance of labeled training data, decreasing their generalizability to orthogonal sources. In this work we propose a novel data augmentation technique that utilizes information from related medical concepts, which improves our model's ability to generalize. Furthermore, we show that incorporating the global context information within the whole medical note (in addition to the traditional local context window), can significantly improve the model's representation for abbreviations. We train our model on a public dataset  (MIMIC III) and test its performance on datasets from different sources (CASI, i2b2). Together, these two techniques boost the accuracy of abbreviation disambiguation by almost 14\% on the CASI dataset and 4\% on i2b2. 
\end{abstract}

\section{Introduction}
\label{sec:intro}

Health care practitioners typically use abbreviations when preparing clinical records, saving time and space with the cost of increased ambiguity. While experienced professionals are usually able to disambiguate abbreviations based on the context, this remains a challenging task for automated clinical note processing. Correctly disambiguating medical abbreviations is important to build comprehensive patient profiles, link clinic notes to ontological concepts, and allow for easier interpretation of unstructured text by fellow practitioners. However, expanding abbreviated terms back to their long-form is nontrivial since abbreviations can have many expansions. For example, ``ra" can mean right atrium, rheumatoid arthritis or room air depending on its context. A number of supervised learning models have been built for abbreviation disambiguation in medical notes, including ones based on Support Vector Machines (SVM) and Naive Bayes classifiers \citep{Moon2012-mf, Moon2013-ow,Wu2017-sn}. However, these methods rely on expensive hand-labelled training data and are vulnerable to overfitting. This is evident in studies where training and testing models on different corpora results in performance drops of 15-40\% \citep{Moon2012-mf, Moon2013-ow,Wu2017-sn, Joopudi2018-bv, Finley2016-vt}. 

The difficulty and cost of creating hand-labelled medical abbreviation datasets is illustrated by the fact that, to the best of our knowledge, there is only one such publically available dataset with training data and labels: CASI \citep{casi}. CASI contains 75 abbreviations, which is just a small fraction of all medical abbreviations. In contrast \citet{AllAcronyms}, a crowd-sourced database that contains abbreviations and their possible expansions, lists \textgreater80,000 medical abbreviations.

\citet{Finley2016-vt} showed that the need for manual annotation can be reduced by reverse substitution (RS). RS auto-generates training data by replacing expansions with their corresponding abbreviations, eliminating the need for manual annotation. They found all the sentences containing possible expansions for an abbreviation in unstructured clinical notes, replaced the expansion with the abbreviated form, and used the expansion as the ground truth label. For example, ``Patient was administered intravenous fluid" becomes ``Patient was given ivf", and the label for the abbreviation ``ivf" is ``intravenous fluid". 

RS, however, creates imbalanced training sets because the distribution of terms in their abbreviated and long forms is often different. For example, the terms ``intravenous fluid" and ``in vitro fertilization" are possible expansions for the abbreviation ``ivf". In notes from the Multiparameter Intelligent Monitoring in Intensive Care (MIMIC-III) \citep{Johnson2016-cu} dataset, the term "intravenous fluid" occurs in its long form 3,132 times, while the term  "in vitro fertilization" does not appear at all. Generating samples for all possible expansions of ``ivf" using RS thus learns a false prior that ``ivf" never expands to ``in vitro fertilization."

\citet{Joopudi2018-bv} improved on standard RS by clustering sentences for abbreviations that performed poorly on their validation set, labelling the centroid of each cluster with the abbreviation’s correct expansion (identified via manual curation), and applying that sense to all sentences in the cluster. Despite improving performance, this method requires hand-labelling, which cannot scale to thousands of abbreviations in datasets such as AllAcronyms. 

An additional problem with medical abbreviation disambiguation is that the local context of a word is not always sufficient to disambiguate its meaning. For example, "rt" could represent "radiation therapy"  or "respiratory therapy", and the phrase "the patient underwent rt to treat the condition" cannot be disambiguated without further information. \citet{Huang} showed that words can be better represented by jointly considering their local and global contexts. \cite{kirchhoff-turner-2016-unsupervised} also demonstrated that document contexts are useful in medical abbreviation disambiguation tasks. A study by \cite{li2015acronym} represented acronyms in scientific abstracts using the embeddings of words with the highest term frequency–inverse document frequency (TF-IDF) weights within a collection of documents. This was motivated by the idea that acronym expansions are related to the topic of the abstract and that topics can be described by words with the highest TF-IDF weights.  

In this work we take a two-pronged approach to improving the accuracy of medical abbreviation disambiguation. First, we demonstrate that we can use prior medical knowledge, in the form of biomedical ontologies such as the Unified Medical Language System \citep{umls} to help create more balanced and more representative examples of training data for RS approaches. Second, we demonstrate that combining local and global context of an abbreviation can help further improve the accuracy of abbreviation disambiguation, achieving 14\% improvements in accuracy on CASI and 4\% accuracy improvements on \citet{i2b2} datasets, all while training exclusively on MIMIC-III data. 

\section{Datasets}
We use five datasets in this study, all of which are publicly available: 

(1) We use clinical notes from MIMIC-III as our training set. We collect sentences from MIMIC III containing abbreviation expansions, as well as concepts in UMLS to augment our training set. We also use MIMIC-III to pretrain word embeddings using FastText and IDF weights.

(2) We augment our training sets based on relationships between expansions and concepts defined by UMLS Metathesaurus.

(3) We use the medical section of AllAcronyms, a crowd-sourced database, to obtain a list of  80,000 medical abbreviations and 200,000 potential expansions. We remove abbreviations that have only one disambiguation and those absent in UMLS, resulting in 30,974 abbreviations.

(4) We validate our method on CASI, a dataset of admission notes, consultation notes, and discharge summaries from hospitals affiliated with the University of Minnesota. After removing abbreviations with one expansion, we had 67 hand-labelled abbreviations with approximately 500 samples per abbreviation. We use this dataset as an orthogonal test set to measure model generalizability.

(5) As another test set we use i2b2, a collection of patient discharge summaries from Harvard Medical School. This dataset does not have hand-labelled annotations, so we use RS to generate labels.

\section{Methods}

\subsection{Overview}

\begin{figure}[h]
  \centering
  \includegraphics[scale=0.5]{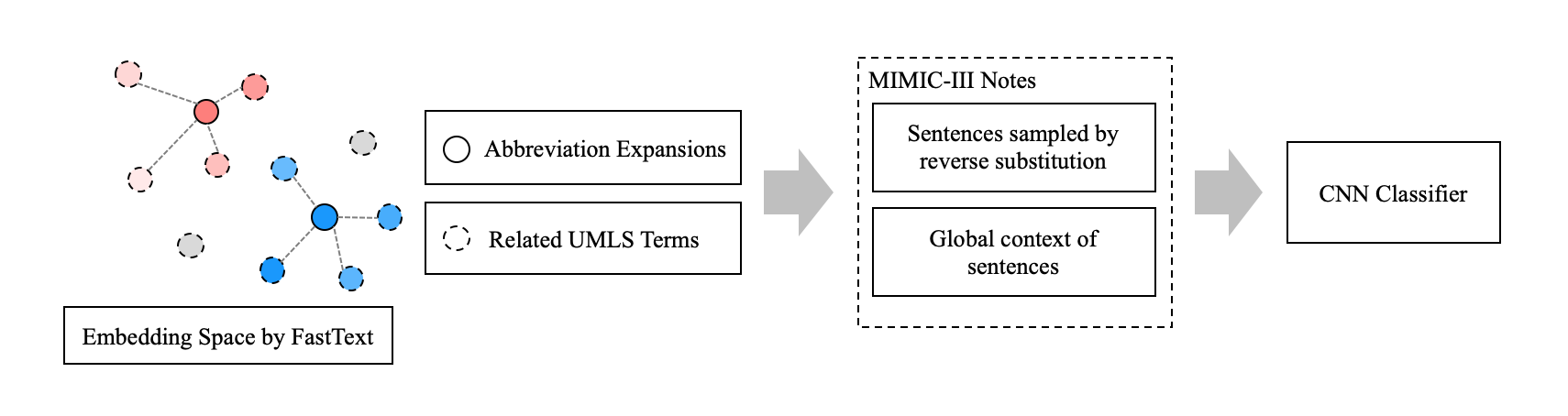}
  \caption{Overview of our method. First, we embed medical concepts from UMLS using a FastText model trained on MIMIC-III clinical notes to map related concepts close in vector space, which we use to augment our training samples. We use MIMIC-III notes to create training samples with RS and generate better embeddings by incorporating global information within the notes. Finally, we train a classifier to perforn the disambiguation task.}
  \label{fig:pipeline}
\end{figure}

An overview of our method is shown in \figureref{fig:pipeline}. While our method follows the overall RS paradigm, in order to reduce the false prior of training sets generated using RS and eliminate the need for labelling abbreviation datasets by hand, we develop a data sampling technique that augments the training set with samples of closely related medical concepts. Our approach is motivated by \citet{Arbabi2019-tj}, who learned embeddings using a medical ontology to identify previously unobserved synonyms for concepts in large unstructured text and modelled concepts missing from the training corpus using ancestors in the ontology. First, we learn word embeddings for terms in clinical notes by training a FastText model on MIMIC-III notes \citep{Bojanowski2017-tc}. We then map medical concepts in UMLS to the resulting vector space to generate a word embedding for every medical concept. Finally, for a given abbreviation, we augment the training samples for each expansion with sentences containing closely related medical concepts determined using embedding distance. This method is easily scalable to previously unseen abbreviations as it does not require any expert annotation.

Using this training set we train a convolutional neural network (CNN) to perform the classification task, which is to predict the correct expansion for an abbreviation given its neighboring words (local context) and the global context of the document, as represented by IDF-weighted word embeddings. 

\subsection{Word embeddings}
To represent input words, we train word embeddings in an unsupervised manner on the MIMIC-III corpus using FastText in order to map semantically similar words close in vector space. The advantage of FastText is that it learns word embeddings by representing each word as a bag of character n-grams \citep{Bojanowski2017-tc}. This is useful for creating good representations of rare words or words not found in the training corpus since we consider sub-word information. We join multi-word medical concepts from UMLS with a “\_” symbol to represent them as a single token. We use this model to embed all medical concepts in UMLS.

\subsection{Training set sampling}
For each expansion for a given abbreviation, we augmented the training samples with the 10 most related medical concepts. We determined the degree of relatedness by measuring the Euclidean distance between the expansion phrase and all concepts in UMLS that are also present in MIMIC-III. We randomly sampled each relative in proportion to its distance from the expansion with replacement according to the following probability:
\[
p_{sampling} = \frac{e^{\frac{-d_r}{T}}}{\Sigma_Re^{\frac{-d_R}{T}}}
\]
where $d_r$ is the Euclidean distance between the expansion and relative and $T$ is the temperature of the distribution. $R$ refers to the 10 closest medical concepts. If sentences for an expansion were present in the training corpus, we treated the expansion as a relative with a distance of $\epsilon$ (a hyperparameter which we set to 0.001).  

We used temperature as a “sharpening” function to change the entropy of the sampling probability \citep{Berthelot2019-dj}. As $T$ approaches 0, the entropy of the distribution decreases and the probabilities approach a one-hot distribution. As $T$ goes to $\infty$, entropy increases and the probability of sampling any relative becomes the same. For each abbreviation, we searched for an optimal temperature value that minimized the loss on the validation set using Bayesian optimization on the MIMIC-III validation set. We constrained the upper and lower bound search spaces for the temperature value to be between $2^{-1}$ and 2, as we found that smaller values overfit to MIMIC-III and reduced generalizability, while larger ones added too much noise through less relevant neighbours.  For each abbreviation, we performed 15 iterations of Bayesian optimization using the Tree-structured Parzen Estimator algorithm \citep{Bergstra2011-vj}. 

\begin{figure}[t]
  \centering
  \includegraphics[scale=0.17]{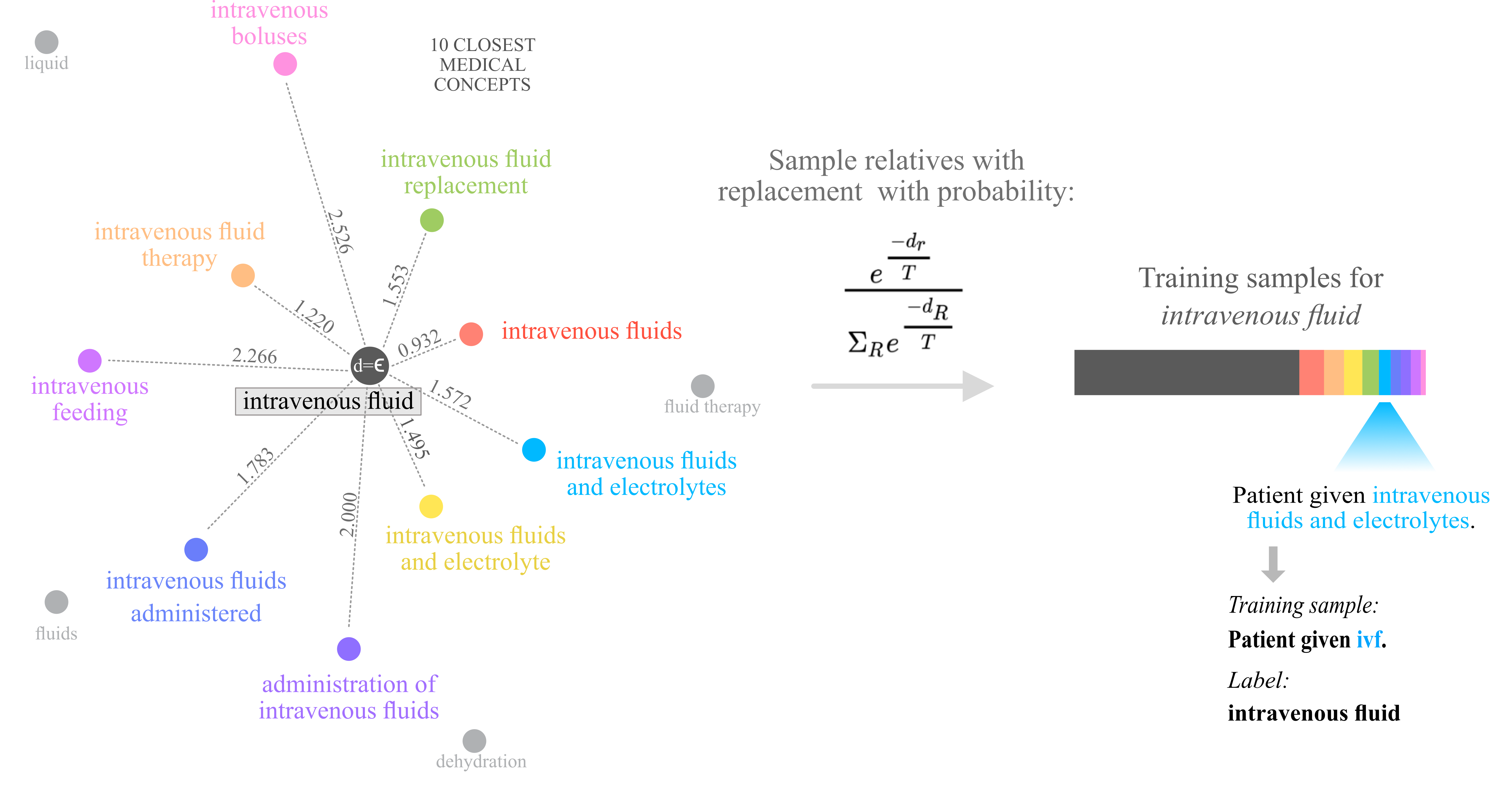}
  \caption{Illustration of data augmentation technique for the training set. For each expansion, we sample sentences for the 10 closest medical concepts using RS proportionally to their Eucledian distance in the embedding space from the expansion. It is shown above the dotted line connecting the expansion to its relative. The probability of sampling is indicated about the arrow. $d_r$ is the Euclidean distance between the expansion and relative and $T$ is the temperature of the distribution. $R$ refers to the 10 closest medical concepts.  In the event that an expansion is present in the training corpus, we sample it with a distance of $\epsilon$, which we set to 0.001. We add the each sample to our training set by replacing the relative with the abbreviation and using the target expansion as the label. An example is this is shown below the colour bar.}
  \label{fig:sampling}
\end{figure}

A schematic of our sampling technique can be viewed in  \figureref{fig:sampling}. As a baseline, we tested our model on the training set only acquired using RS (i.e. it was not augmented using related medical concepts). As a second baseline, we tested our model on the training set that was sampled with replacement, in that the expansions were sampled with replacement so that we had an equivalent number of training samples per expansion. This was to ensure that any change in performance could only be attributable to incorporating auxiliary medical knowledge, and not unbalanced training datasets from rare abbreviations.

\subsection{Sentence embeddings}

We map an input sentence to a vector representation using a simple encoder similar to the one used by  \citet{Arbabi2019-tj}. The network consists of one convolution layer with a filter size of one word, followed by ELU activation \citep{clevert2015fast}. Max-pooling-over-time pooling is used to combine the output into a single vector, $\vec{v}$: 
\[
\vec{v} = max_t(ELU(\vec{W_1}\vec{x}^{(t)} + \vec{b}))
\]
where $\vec{x}^{(t)}$ is the word embedding of the term at index \textit{t}. $\vec{W_1}$ and $\vec{b}$ correspond to the weight matrices and bias vectors, respectively, which we learn through training.

A fully connected layer with ReLU activation followed by L2 normalization is used to map \textbf{v} to the final encoded sentence representation:

\[
\vec{e} = \frac{ReLU(\vec{W_2}\vec{v})}{||ReLU(\vec{W_2}\vec{v})||_2}
\]

The embedded sentence is a representation of the local context. To incorporate the global context of a sample, $\vec{g}$, we take the weighted average of the embedding vectors for each word in the document. The embeddings are weighted using IDF weights trained on the MIMIC-III corpus. The vector $\vec{g}$ is calculated as follows:

\[
\vec{g} = \frac{\sum_{i=1}^{d} \vec{u_{i}}*w(t_{i})}{\sum_{i=1}^{d} w(t_{i})}, i \neq j
\]
where \textit{j} is the index of the abbreviation, \textit{i} is the index of the \textit{i}-th word in the document, and \textit{d} is the number of words in the document; $\vec{u_i}$ is the word embedding and \textit{w(t\textsubscript{i})} is the IDF-weighting of the \textit{i}-th word.

We then concatenate $\vec{g}$ with the encoded sentence vector, $\vec{v}$ and normalize it to produce the final encoded sample embedding:

\[
\vec{e} = \frac{ReLU(\vec{W_2}\vec{[v;g]})}{||ReLU(\vec{W_2}\vec{[v;g]})||_2}
\]

\subsection{Classification using a convolution neural network}

\begin{figure}[t]
  \centering
  \includegraphics[scale=0.4]{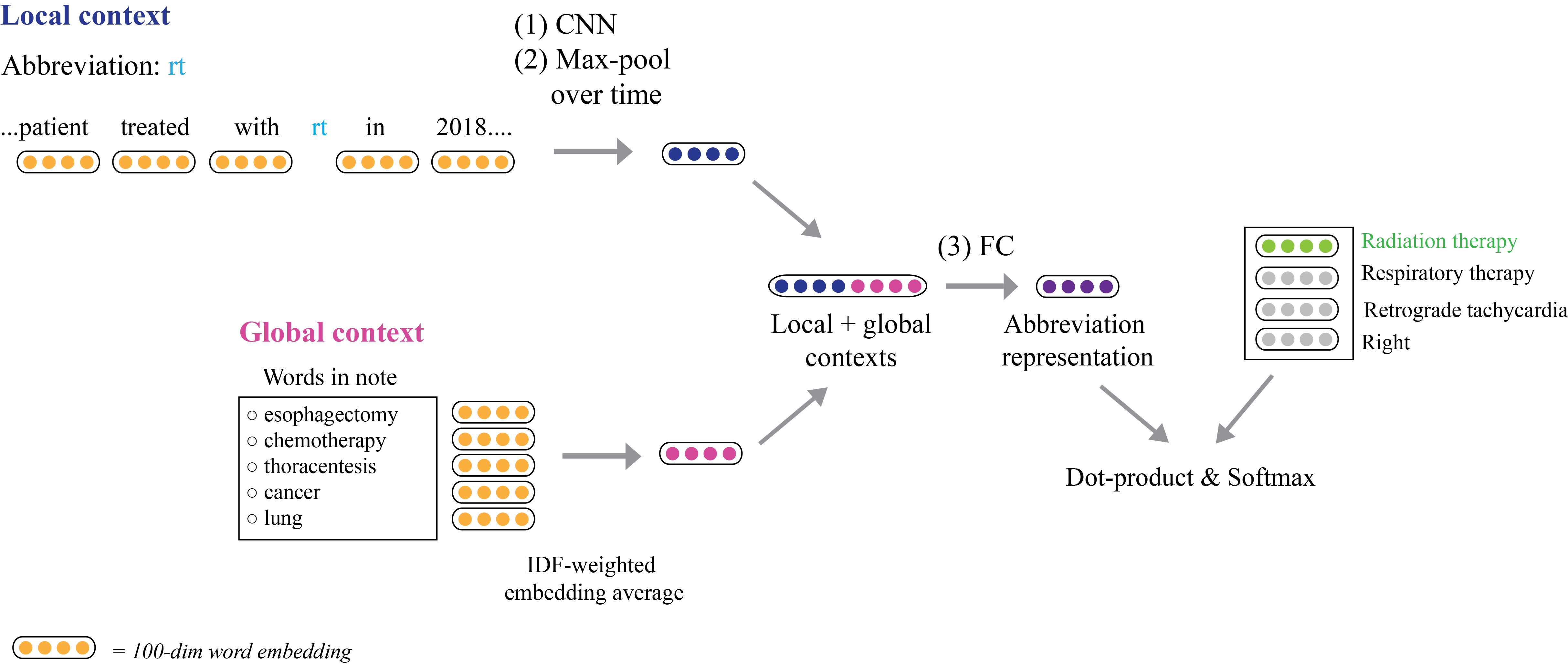}
  \caption{Overview of our abbreviation disambiguation model. Sentences containing a target concept are passed through a convolutional neural network (CNN) and max-pooled over time to generate an encoding of the local context. Global context takes the IDF-weight average of word embeddings in the entire document. We combine global context with the output from the sentence encoder and pass it through a fully-connected layer (FC). We maximize the dot-product of the encoded sentence and expansion embedding.}
  \label{fig:model}
\end{figure}

Our model is trained to minimize the distance between a target expansion embedding and its context (Figure ~\ref{fig:model}). 
Our model represents expansion embeddings with an embedding matrix, $\vec{H}$, where each row, $\vec{H}_c$, corresponds to the embedding of an expansion for a given abbreviation. To do the
classification task of assigning an expansion label, $c$, to an input sentence, $\vec{e}$, we take the dot-product of $\vec{H}$ and $\vec{e}$ and apply a softmax function, such that:

\[
p(c|\vec{e}) = \frac{exp(\vec{H}_{c}\vec{e})}{\Sigma_{c'}exp(\vec{H}_{c'}\vec{e})}
\]

We label the abbreviation with the expansion having the largest probability $p(c|\vec{e})$.

\section{Experiments}

We trained our model on sentences from MIMIC-III. We collect sentences containing expansions from CASI and medical concepts from UMLS using RS. In total, 105,161 concepts in UMLS are found in MIMIC-III. To learn word vectors, we trained the FastText model as described in \citet{Bojanowski2017-tc}.  For the classification task, we built one model for each abbreviation. To train our model, we used a maximum of 500 samples per expansion and a context window of 8 words. On average, each abbreviation has 3.46 expansions. We train our models on 60\% of the sample set, validate it on 20\%,  and keep 20\% as a held-out test set. We train all concept embedding models for 100 epochs with a learning rate of 0.01. We use early stopping on the validation loss to prevent overfitting. 

We consider two forms of accuracy: Micro accuracy is the total number of abbreviations correctly disambiguated divided by the total number of samples in the test set across all abbreviations with two or more possible expansions. Macro accuracy is the average of individual abbreviation accuracies.

We report the performance of our classifier on three different training sets. The first training set (Control) consists of samples solely acquired using RS without any alterations. The second (SWR) is similar to the first training set, except that we sample training sentences with replacement such that each expansion has an equivalent number of training samples. The third training set (Full) incorporates our novel data sampling technique by including medical relatives of expansions into the training set. We sample concepts with replacement so that all expansions have an equivalent number of training samples. We also train each model both using only local neighborhood of the abbreviation and with incorporating global context information for each sample (+ global). We use bootstrapping to obtain the mean for each abbreviation by resampling our predicted values and true values 999 times. A Wilcoxon signed rank test was used to compare the macro accuracy results of different models (micro accuracy is a point estimate). We evaluated our model on three datasets: a held-out test set consisting of RS samples of abbreviation expansions from MIMIC-III, an orthogonal dataset of 67 abbreviations from CASI with gold-standard annotations and 403 abbreviations from i2b2 generated. We generated the i2b2 samples by finding sentences with expansions from AllAcronyms using RS.  

\section{Results}

\begin{figure}
  \centering
  \includegraphics[scale=0.12]{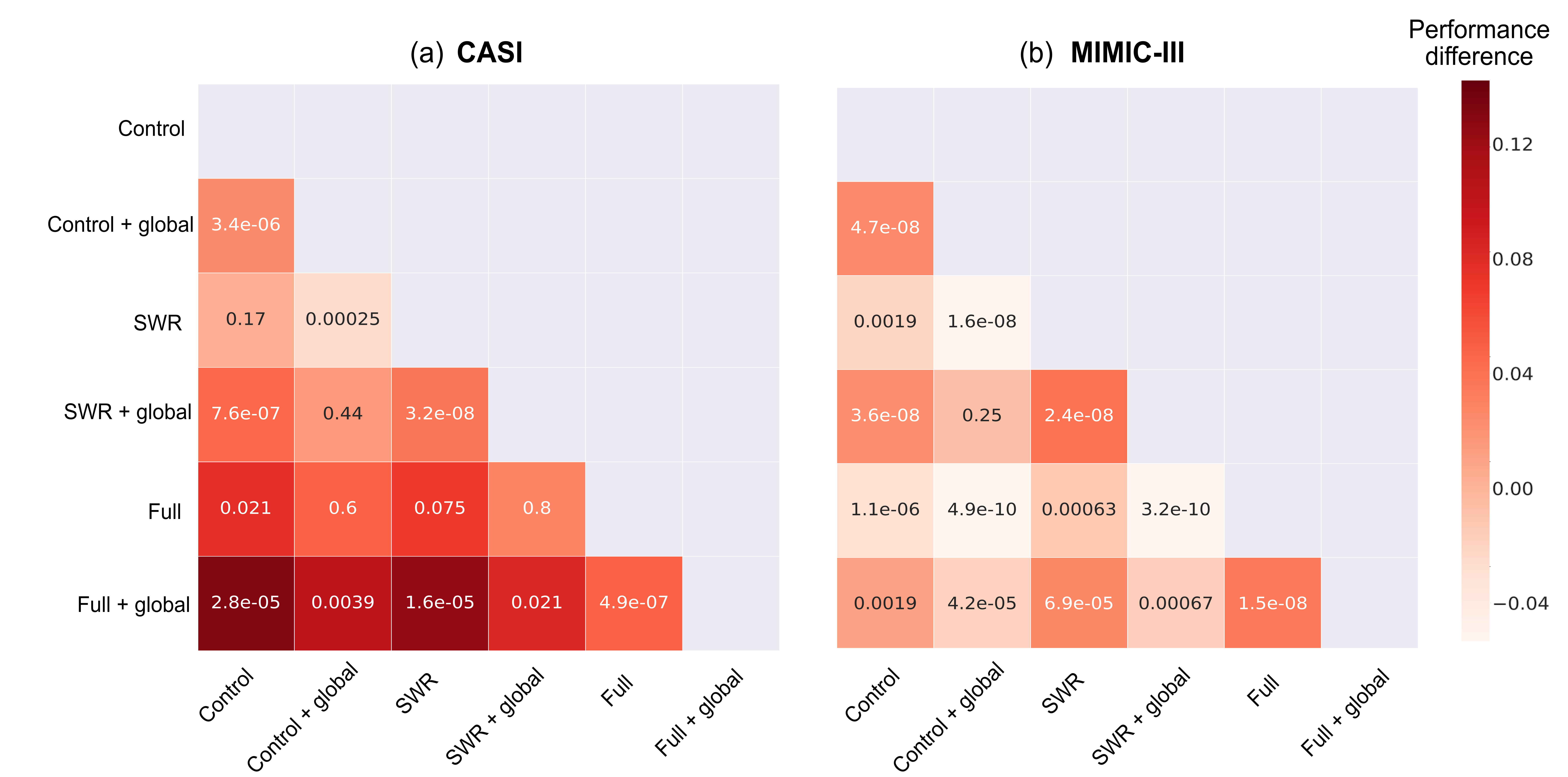}
  \caption{Matrix showing performance differences and p-values between all models on (a) CASI and (b) MIMIC-III test sets. The colour intensity of each square reflects the performance difference between the corresponding model on the vertical axis and model on the horizontal axis. p-values were obtained using a Wilcoxon signed rank test and are displayed inside each square.}
  \label{fig:matrix}
\end{figure}

\tableref{tab:tab1} shows the micro and macro accuracies of our concept embedding model using our data augmentation technique on test sets from MIMIC-III and CASI.  p-values and performance differences between all models are displayed in \figureref{fig:matrix}. We find that training abbreviation with both local and global contexts gives significantly better performance than training on local alone. We also find that augmenting the training set with related medical concepts marginally decreases the performance of our model when tested on MIMIC-III. This was expected, as we are augmenting the data with noisy labels, and the abbreviations that we are now better able to predict do not actually appear in MIMIC-III. However, this makes the model more generalizable to orthogonal datasets, as there is an 8\% (p=0.02) increase in accuracy on CASI compared to the control. Incorporating global context increased this value to 14\% (p=5e-07). Notably, the improvement achieved on CASI by adding global context grew, from 3\% when using the control model to 5.5\% when using the full model. This demonstrates that the global context in which related terms appeared aided disambiguation, even if the local context may have been different. 

\begin{figure}
  \centering
  \includegraphics[scale=0.20]{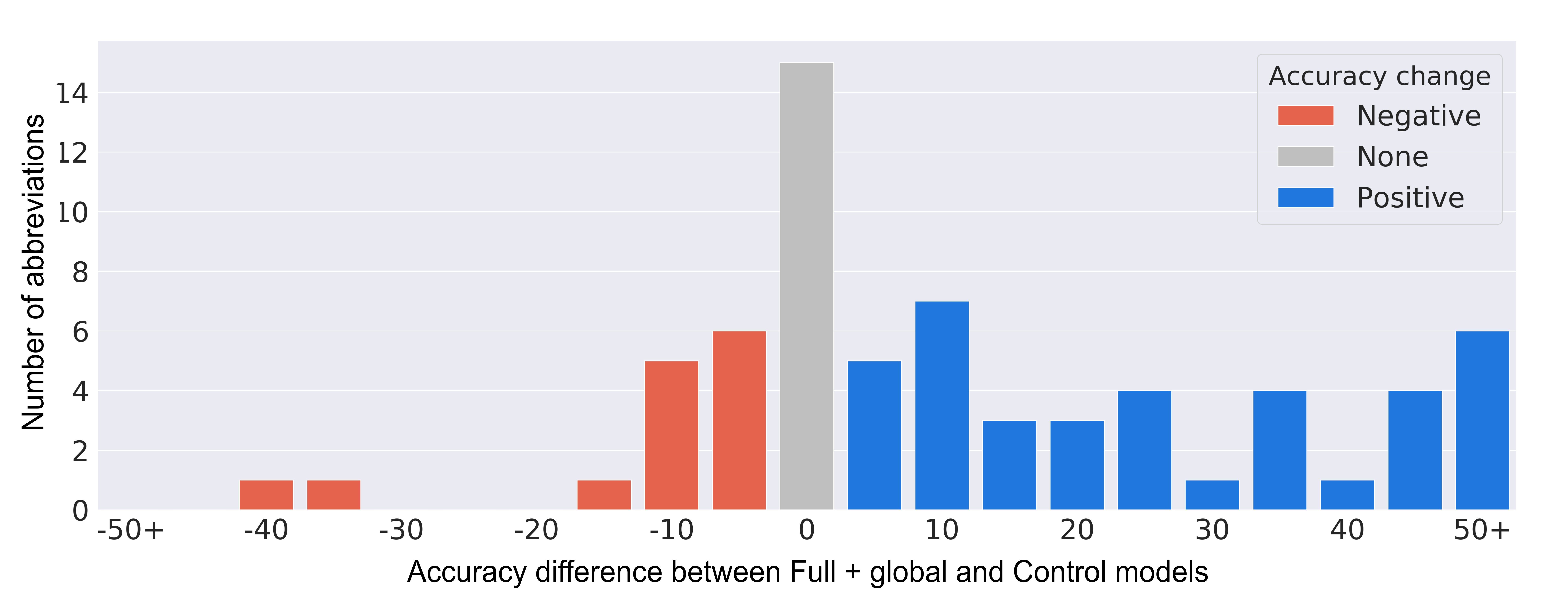}
  \caption{Histogram showing the accuracy difference between the best model and control model performance (\%) on CASI abbreviations. The x-axis shows buckets of 5\%, where each unit is the bucket mean (i.e. x=0 buckets data from an accuracy difference of -2.5\% to +2.5\%). }
  \label{fig:histogram}
\end{figure}

\figureref{fig:histogram} is a histogram displaying the performance difference between our best model (Full + global) and the control model for CASI abbreviations. Notably, the performance improved for 38 abbreviations. The abbreviations that benefited most from our model were the ones where an expansion did not appear in the training corpus or appeared at a very low frequencies. For example, our model increased the performance for the abbreviation ``na" by 75\% compared to the control. This is because the phrase ``narcotics anonymous", a possible expansion for ``na", only appears twice in MIMIC III. Upsampling that phrase and incorporating related concepts such as ``alcoholics anonymous" and ``nicotine use" enabled us to create a better representation for it. 

\begin{table}
  \small
  \caption{Micro and macro accuracy of (\%) of our model on 67 CASI abbreviations trained on data generated using RS (control), RS where we sample training data with replacement (SWR), and RS with replacement and augmentation with related medical concepts (Full). For CASI we use the full dataset as testing. For MIMIC we use the 20\% held out fraction with RS labels.}
  \label{sample-table}
  \centering
  \begin{tabular}{p{3.5cm}llll}
    \toprule
    \multirow{2}{*}{Sampling Method} & \multicolumn{2}{c}{MIMIC TEST} &
    \multicolumn{2}{c}{CASI TEST} \\
    \cmidrule(r){2-3}
    \cmidrule(r){4-5}
         & Macro Accuracy     & Micro Accuracy & Macro Accuracy     & Micro Accuracy \\
    \midrule
    Control & 0.913  & 0.866 & 0.621  &  0.625  \\
    Control + global  & \bf{0.944}  & \bf{0.917} & 0.651 & 0.654   \\
    SWR     &0.897 & 0.854  & 0.629  & 0.631   \\
    SWR + global     &0.942 & 0.912  & 0.671 & 0.674   \\
    \midrule
    Full     & 0.888       & 0.836 & 0.705* & 0.704  \\
    Full + global   & 0.929      & 0.899 & \bf{0.760}** & \bf{0.760}  \\
    \bottomrule
  \end{tabular}
    \begin{tablenotes}
    \item [*] *p\textless0.03 (Wilcoxon signed rank test compared with Control model)
    \item [**] **p\textless5e-7 (Wilcoxon signed rank test compared with Full model) 
  \end{tablenotes}
  \label{tab:tab1}
\end{table}

\tableref{tab:tab2} shows the performance of our model tested on a larger test set of 403 abbreviations (chosen based on lexicographic order), using another orthogonal dataset, i2b2, with labels generated using RS. While the full model still outperforms the control, the performance gain is more modest (4\%).  The smaller improvement on i2b2 may be indicative that this dataset more closely resembles MIMIC, in terms of the frequency of different disambiguations. For example in the case of ``ivf", there are significantly fewer examples of fully spelled out cases of "in vitro fertilization" than "intravenous fluids" in both MIMIC (zero versus 2503) and i2b2 (2 versus 49). At the same time "in vitro fertilization" is the more common expansion in CASI (294 versus 181). This could be indicative either in a difference between the datasets, or human behavior: the RS method relies on the long form of an abbreviation to be written out fully, and this may be less likely with abbreviations that are either clearer in the context, or are longer.

\begin{table}
  \caption{Micro and macro accuracy of (\%) on 403 i2b2 abbreviations trained on data generated using RS (control) and our best model from Table 1 (Full + global).}
  \label{sample-table}
  \centering
  \begin{tabular}{p{3.5cm}llll}
    \toprule
    & \multicolumn{2}{c}{i2b2 TEST} \\
    \cmidrule(r){2-3}
    \cmidrule(r){4-5}
    Method     &  Macro Accuracy     & Micro Accuracy \\
    \midrule
    Control  & 0.689  &  0.521  \\
    Full + global  & 0.729* & 0.577  \\
    \bottomrule
  \end{tabular}
      \begin{tablenotes}
    \item [*] *p=6e-11 (Wilcoxon signed rank test compared with Control model)
  \end{tablenotes}
  \label{tab:tab2}
\end{table}

\section{Conclusion}

Our contributions in this paper are twofold. First, we demonstrate the usefulness of prior medical knowledge, in particular the UMLS ontology, to develop a novel data sampling technique that creates good representations for abbreviations that are missing or infrequent in the training corpus. For all samples we are also able to generate better representations by considering the global context in which an abbreviation appears. Because of these improvements, our overall framework demonstrates 14\% higher accuracy of  abbreviation disambiguation on the auxiliary CASI dataset with hand-labelled abbreviations.

Another advantage of our method over previous work is that it can scale to thousands of abbreviations as it requires no hand labelling, which we demonstrate by utilizing it on both MIMIC (training/testing) and i2b2 (orthogonal testing) datasets for 403 abbreviations,  showing 4\% accuracy improvement relative to control models. 

\acks{The authors would like to thank Erik Drysdale, Nicole Sultanum, and Devin Singh for insightful discussions, as well as members at the Centre for Computational Medicine (CCM) for technical support, especially Pouria Mashouri and Rob Naccarato.}

\bibliography{abbrdisamb_neurips_ml4h_2019_cameraready}

\end{document}